\newcommand{\cv}[1]{}  %
\newcommand{\av}[1]{#1} %
\av{

\documentclass[10pt,usletter]{article}
\usepackage{geometry}
\usepackage{amssymb} 
\usepackage{mycitations}  %
\newcommand{\citet}[1]{\citeauthor{#1}~[\shortcite{#1}]}

\geometry{
    letterpaper,
    width=5.5in,
    height=8.5in,
    top=1in,
    headheight=14pt
}

  }

\usepackage{times}
\usepackage{soul}
\usepackage{url}
\usepackage[hidelinks]{hyperref}
\usepackage[utf8]{inputenc}
\usepackage[small]{caption}
\usepackage{graphicx}
\usepackage{amsmath}
\usepackage{amsthm}
\usepackage{booktabs}
\usepackage{algorithm}
\usepackage{algorithmic}
\usepackage[switch]{lineno}
\usepackage{subcaption}
\usepackage{listings}
\usepackage[most]{tcolorbox}

\cv{\linenumbers}

\title{Extracting Problem Structure with LLMs for\\ Optimized SAT Local
  Search\thanks{Research supported by the Austrian Science Fund (FWF)
within the projects  10.55776/P36420 and
10.55776/COE12.}
}

\cv{
\author{
    Anonymous
}

}

\av{
\author{Andr\'{e} Schidler and Stefan Szeider\\[4pt]
  \small  Algorithms and Complexity Group\\[-3pt]
  \small TU Wien, Vienna, Austria\\[-3pt]
\small \texttt{[aschidler|sz]@ac.tuwien.ac.at}
}
\date{}
}

\usepackage{xcolor}
\newcommand{\Card}[1]{|#1|}
\colorlet{MyRed}{red!50!black!100!}

\begin{document}
\maketitle
\av{\thispagestyle{empty}}
\begin{abstract}
  Local search preprocessing makes Conflict-Driven Clause Learning
  (CDCL) solvers faster by providing high-quality starting points and
  modern SAT solvers have incorporated this technique into their
  preprocessing steps. However, these tools rely on basic strategies
  that miss the structural patterns in problems. We present a method
  that applies Large Language Models (LLMs) to analyze Python-based
  encoding code. This reveals hidden structural patterns in how
  problems convert into SAT. Our method automatically generates
  specialized local search algorithms that find these patterns and use
  them to create strong initial assignments. This works for any
  problem instance from the same encoding type. Our tests show
  encouraging results, achieving faster solving times compared to
  baseline preprocessing systems.\end{abstract}

\section{Introduction}
Local search preprocessing guides CDCL solvers to faster solutions through better starting points. \citet{BalintM13} showed that preprocessing improves CDCL-based SAT-solving performance. Current SAT solvers like CaDiCaL \cite{Biere19,BiereFHH20,Biere24} and CryptoMiniSat \cite{Soos20} have adopted this approach in their preprocessing phase. These tools apply basic strategies that work well for random problems but miss critical patterns in structured instances.
SAT encodings of real problems contain inherited patterns from graph layouts, data connections, and domain-specific rules. The transformation to Conjunctive Normal Form (CNF) obscures these patterns. Current local search methods skip these structures in favor of general approaches.
This paper addresses these limitations by introducing a framework that leverages LLMs to generate local search strategies tailored to encoding structures, enabling solvers to take advantage of these patterns for improved performance.
Our research addresses three questions:
\begin{enumerate}
\item How can LLMs analyze PySAT \cite{IgnatievTK24} code to interpret
  how problem structure translates to SAT clauses?
\item How can we create local search strategies that recognize and
  exploit these encoding patterns?
\item What performance gains do structure-aware preprocessing methods
  achieve versus standard approaches?
\end{enumerate}
Our method applies LLMs to read and interpret the PySAT code that
converts structured problems to SAT. The model spots high-level
constructs like graph connections, path constraints, or counting
limits and then builds these insights into a specialized local search
procedure. This procedure targets the encoding rather than specific
instances, making it applicable across all instances encoded this way.
This use of LLMs moves beyond generic applications of AI tools by
integrating them directly into the SAT solving workflow.
While existing research highlights using LLMs for interactive
code generation, such as in GitHub Copilot \cite{JiangWSKK2024}, our
approach addresses the challenges associated with fully automated code
generation for specialized tasks like SAT preprocessing.
Significantly, the strategies we generate are encoding-specific but
apply across problem instances, addressing a key gap in SAT-solving
techniques.

An automated pipeline tests and corrects each local search
procedure.  The framework systematically generates, validates, and
refines a diverse set of local search strategies through a two-phase
process: a gathering phase to encourage diversity and a refinement
phase to improve quality. This method automates aspects of algorithm
design while maintaining performance guarantees, reducing the reliance
on manual, domain-specific effort.

We evaluated our method on multiple structured problems, including
Directed Feedback Vertex Set, Bounded Depth Decision Trees, and
Treewidth. Our results demonstrate the effectiveness of using LLMs to
generate encoding-specific preprocessing algorithms. While challenges
remain in adapting these algorithms to encoding requirements, the
approach provides a reproducible solution for improving SAT
preprocessing through automation. The LLM-generated algorithms
effectively exploit encoding structure, with several variants
outperforming baselines on hard instances. For Directed Feedback
Vertex Set specifically, our methods solved 12 additional instances
beyond conventional SAT solvers while maintaining performance on
easier cases, demonstrating the potential for LLMs to enhance SAT
solving through automated, problem-aware preprocessing.

\av{\paragraph{Supplementary Material} Code and instances are available on
  Zenodo \cite{SchidlerSzeider25}}

\section{Background and Related Work}

\av{\subsection{SAT solving}}
The propositional satisfiability problem (SAT) takes a propositional
formula and asks if there exists a variable assignment that makes the
formula true, i.e., that \emph{satisfies} the formula.
Modern complete SAT solvers use Conflict-Driven Clause Learning (CDCL)~\cite{SilvaS96,DBLP:series/faia/0001LM21,FichteBHS23}
and search for a satisfying assignment or proves that no such assignment exists.

A common way of using SAT solvers is by \emph{encoding} other
problems into SAT, i.e., representing
 a problem instance in propositional logic such that the satisfiability/unsatisfiability
of the formula corresponds to a Yes/No answer for the original problem; and a satisfying
variable assignment corresponds to a solution of the original problem.
An example would be encoding whether a given graph admits a
proper $k$-coloring, where
a satisfying assignment can be translated into a $k$-coloring.
An encoding requires an \emph{encoding scheme} that translates instances from the original
problem into propositional logic and thereby creates the actual encoding of an instance.
In our example, it would be an algorithm that produces for a given
graph and integer $k$ an encoding that is satisfiable if and only if
the graph admits a $k$-coloring.

\cv{\paragraph{Local Search Algorithms for SAT}}
\av{\subsection{Local Search Algorithms for SAT}}
The GSAT algorithm \cite{SelmanLM92} introduced the basic local search
approach for SAT. GSAT picks the variable flip that leads to the most
satisfied clauses. WalkSAT \cite{SelmanKC94}
operates by randomly selecting an unsatisfied clause and then choosing a variable within that clause to flip. %
These algorithms sparked numerous variations:
Novelty \cite{McAllesterSK97} considers the time since a variable's last flip when selecting moves. Adaptive Novelty+ \cite{Hoos02} automatically tunes its noise parameter during search. ProbSAT \cite{BalintS12} uses probability distributions based on make and break values to choose variables. YalSAT \cite{Biere14} combines ideas from several algorithms with additional restart strategies.

\cv{\paragraph{Local Search in CDCL Solvers}}
\av{\subsection{Local Search in CDCL Solvers} }
Modern CDCL solvers use local search during preprocessing to find promising
initial assignments. CaDiCaL \cite{Biere19,CaiZFB22,Biere24} implements a variant of
ProbSAT in its rephasing procedure.
The solver stores good assignments found during search and uses them as starting
points after restarts. This hybrid approach performs particularly well
on random and hard combinatorial instances by combining the systematic nature of CDCL
with local search's ability to find solutions in satisfiable regions quickly.
The integration of local search into CDCL brings two main benefits:
better initial assignments can guide the solver toward solutions faster,
and local search patterns can inform restart strategies.
However, current implementations use generic local search methods
that don't account for problem structure.

\cv{\paragraph{Machine Learning and SAT solving}}
\av{\subsection{Machine Learning and SAT solving}}
Machine learning has enhanced SAT solving in multiple ways. Deep neural networks can learn variable selection policies that speed up SAT solving \cite{SelsamB19}.  A recent line of work leverages graph neural networks  to guide SAT local search \cite{YolcuP19}.
LLMs have recently shown success in algorithm generation. AlphaCode
\cite{LiCCKSLEKGLHCMBCHWGCMCMRKFKV22} and CodeGen
\cite{NijkampPHTWZSX23} can generate correct
implementations of algorithms.
\av{For constraint satisfaction, system
StreamLLM generates streamlining constraints using LLM-calls
\cite{StreamLLM,VoborilRS24}, and MCP-Solver, on the other hand, allows
interactive calls to a constraint solver within an LLM chat via the
Model Context Protocol \cite{Szeider24}.}
However, using LLMs to analyze and
improve algorithms is still a new direction. Related to our approach
is work that uses LLMs to optimize compiler passes \cite{CumminsSGELRGGHSL23}.
For our target problems, SAT encodings remain competitive with
specialized algorithms.

The work in this paper differs from this related work in
three fundamental ways:
(i) we use LLMs to generate specialized local search algorithms, not general SAT solvers,
(ii) we focus on problem structure, not instance-specific features, and
(iii) we provide runtime and correctness guarantees.

\section{Problem Statement}

We consider the problem of automatically finding problem specific local search
approaches that perform well in conjunction with CDCL SAT solvers.
Existing local search approaches have two drawbacks.
First,
these
algorithms are usually general purpose algorithms that do not
consider, or know of, the encoding scheme that has been used to create
the formula.  We expect that special considerations for the encoding
scheme boosts the performance of local search.
Second,
it has been shown empirically that local search often struggles to
find satisfying assignments on its own~\cite{LiL12,CaiZ21,CaiZFB22}.
As discussed in the previous section, hybrid approaches that
combine local search and CDCL solvers often achieve better results on
hard combinatorial instances than either paradigm on its
own~\cite{CaiZFB22}.
Hence, local search methods specifically designed for hybrid approaches
that are specific to the encoding scheme promise an improved
performance on these hard instances.

Creating
specialized algorithms is a time-consuming effort and is often focused
on well-known approaches.
This focus is necessary, as manually creating specialized algorithms with a
variety of approaches is usually infeasible.
Hence, we evaluate if automatically
generating prototypes using large language models (LLMs) is feasible,
such that these prototypes implement a wide variety of different approaches.

We evaluate the performance of a local search function
by setting the \emph{default phases} of the CDCL solver
to the assignment found by the local search function.
The default phase of a variable is either true or false and determines
which value the CDCL solver assigns the variable whenever the solver
chooses a value for the variable.
In the best case, the local search finds a satisfying assignment, in which
case the CDCL solver will terminate almost immediately.
Further, the closer the local search assignment is to a satisfying assignment,
the faster the CDCL solver is expected to solve the instance.
We then measure how long the SAT solver takes to solve an instance.
This can be compared against the SAT solver using no local search or
other local search functions.

We apply our approach to three carefully chosen combinatorial
problems with increasing encoding complexity: Graph Coloring with its
straightforward CNF representation, Directed Feedback Vertex Set with
its more sophisticated reachability-based encoding, and Bounded Depth
Decision Trees with its highly complex encoding that captures
intricate decision paths. This selection allows us to systematically
explore the capabilities and limitations of our approach as the
underlying SAT encodings become more intricate.

\paragraph{Graph Coloring (Coloring)}
Coloring is one of Karp's original 21 NP-complete problems~\cite{karp1972reducibility}.
Given a graph $G = (V, E)$ and a positive integer $k$, we need to assign each vertex
one of $k$ different colors such that
adjacent vertices are not monochromatic.
The classic encoding scheme is comparatively simple~\cite{Gelder08},
checking whether $G$ permits a $k$-coloring
requires
$O(\Card{V} \cdot k)$
many clauses.

\paragraph{Directed Feedback Vertex Set (DFVS)}
Given a directed graph $G = (V, A)$ and a non-negative integer $k$,
the goal is to find a set of vertices with cardinality at-most $k$,
such that after removing these vertices,
the directed graph is acyclic.
There are different ways to encode this into SAT~\cite{JanotaGM17}.
We use an encoding scheme that encodes reachability, which requires $O(\Card{V}^3)$
many clauses.
Further, additional complexity is introduced by the cardinality constraint used
to limit the number of removed vertices.

\paragraph{Bounded Depth Decision Trees (BDDT)}
Given a labeled dataset with numerical features and a non-negative integer $k$,
the goal is to find a decision tree with depth at-most $k$ that correctly
predicts the label (or class) for the whole dataset.
We use the encoding scheme by \citet{ShatiCM21}, which is fairly
complicated as it has to encode, among other things, every possible path
through a tree of depth $k$.

Next, we discuss our approach for generating these local search functions.

\section{Methodology}
\begin{figure}
    \centering
\cv{    \includegraphics[width=\linewidth]{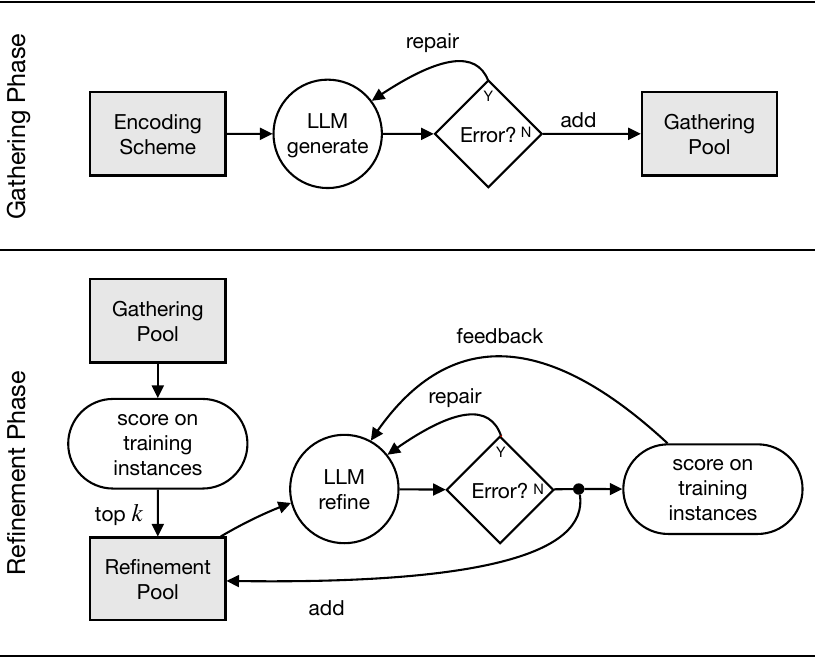}}
\av{\includegraphics[width=.7\linewidth]{map-2}}
    \caption{Schematic of the whole approach from the encoding scheme
      to pool of refined local search functions. From this pool we
      take the top-performing functions for the evaluation on test
      instances.}
    \label{fig:pipeline}
\end{figure}

In this section, we describe our approach that is sketched in Figure~\ref{fig:pipeline}.
The process consists of two phases,
a \emph{Gathering Phase} and a \emph{Refinement Phase}.
We will describe each part of the process, starting with the input,
the encoding scheme.

\subsection{Standardizing Encoding Schemes}
There are many ways to implement encoding schemes.
Since we are defining a general approach for generating local search methods, we need a
standardized way of representing encoding schemes.
We express our encoding schemes in Python using the popular framework
PySAT~\cite{Ignatiev18}.  With PySAT, one can express the encoding scheme in
terms of understandable---both by a human and an LLM---programming
constructs.  Further, PySAT abstracts away cardinality constraints
behind a single function.

\subsection{Gathering Phase}
We provide the LLM with the PySAT encoding scheme,
without stating the problem it encodes or any information on the instances
we will test the local searches on.
The LLM is then requested to return a local search function that meets the following specifications:\footnote{
The prompts are given in the supplementary material.
}
\begin{itemize}
    \item Performs local search for the encoding.
    \item Has a specific function name and takes as input the original instance,
        the encoded instance as PySAT objects, and a timeout.
    \item Returns a (partial) variable assignment where a Boolean value can be accessed using the variable identifier.
    \item Must return within the specified timeout.
    \item Is encouraged to use the input instance and the structure of the encoding,
    \item as well as novel approaches.
\end{itemize}

Whenever we receive a local search function, we verify its correctness (``Error?''
 in Figure~\ref{fig:pipeline}).
We verify that the local search function can be executed, finishes within the timeout,
and returns a (partial) assignment by running the
local search function on an easy instance for half a minute.
Whenever an error is detected, the LLM is provided the error and,
when applicable, the line the error occurred in.
The LLM is then asked to repair the issue.
This repair cycle is repeated, if necessary, until either the code passes the verification,
or a defined limit is reached.

Every local search function we find is added to the prompt's context, such that
the LLM does not repeat it or similar functions.
We use an increasing temperature to encourage increasingly creative ideas.

\subsection{Scoring}\label{sec:score}
Since we aim to generate a wide variety of local search functions,
we require a method to automatically compare their performance.
Given a local search function, we run it on a set of training instances
(``score on training instances'' in Figure~\ref{fig:pipeline}).
For each instance we let the local search find an assignment and pass it
to a SAT solver that tries to solve the instance based on the provided assignment.
This allows us to rank a set of local search functions.
All local search functions that
caused a runtime error are ranked last and we use the following
criteria for the remaining functions, lower values are better.
\begin{enumerate}
    \item The number of instances where the local search function did not return an
        assignment within the timeout.
    \item The number of instances with a SAT solver timeout.
    \item The average runtime over successful SAT solver calls.
\end{enumerate}

\subsection{Refinement Phase}
In this phase, we focus on improving the existing local search functions
found in the previous phase.
 We pick the top searches from the gathering phase
and process them one by one.
Therefore, refinement runs are independent of each other.
 We start from the original request and search function as context and
 ask the LLM to vary the function, while not changing the overall
 idea.
Each version is scored as described above and the LLM
receives feedback about the function's relative performance
to the previous version, where a SAT runtime change
is only considered significant if it the average runtime changed by
more than 10\%.
Depending on the function's performance being better or worse,
the LLM is requested to either continue with similar refinements,
or revert and try a different approach.
In case there is no significant difference, the LLM
is prompted to perform a bigger change.

After completing the refinement phase,
we pick the top local search functions for our final Test Evaluation.

\section{Experimental Evaluation}

In this section, we evaluate the quality the generated local search functions.
Further, we also explore how well the different parts of our approach
contribute to this quality.\footnote{Code and Results are in the Supplementary Materials.}

\subsection{Setup}
We use the OpenAI models o1-mini-2024-09-12 and gpt-4o-2024-11-20, as well as Anthrophic
claude-3-5-sonnet-20241022.
We run gathering and refinement on a MacBook M1 and
the Test Evaluation on servers with two AMD EPYC 7402 CPUs having 24 cores running
at 2.80GHz.
Each run has a memory limit of 128 GB.
We use Cadical~1.9.5 as the SAT solver and PySAT~1.8.dev13.

We consider one separate set of benchmarks for each of our
three encoding schemes.
Since we encode optimization problems,
we require an upper bound for the encoding, e.g., we need some good $k$ to encode
the decision problem, if a graph allows a $k$-coloring.
We find an upper bound differently for each problem and describe
the procedures subsequently.
We split the benchmarks for each problem instances into two sets.
\emph{training instances} are instances
solved by the SAT solver alone in between 10 and 60 seconds
and are used to quickly determine the quality of local search functions.
\emph{test instances} are the instances that take longer than one minute to solve.

\paragraph{Graph Coloring}
We consider 257 graphs from TreewidthLib\footnote{The graphs were kindly
provided by \citet{Fichte17}.} and
DIMACS\footnote{\url{https://sites.cc.gatech.edu/dimacs10/}}.
For each instance,
we establish an upper bound using DSATUR~\cite{Brelaz79}
and then improve this bound for ten hours using
the SAT encoding.
Further, whenever we know a better upper bound from literature~\cite{SunHZL21},
we decrease the upper bound by one to obtain a hard instance.
This results in 10 training and 38 testing instances.

\paragraph{DFVS}
\av{\sloppypar} We follow the evaluation in~\cite{KieselS23} and use 513 graphs
from PACE~2022~\cite{PACE22}, ICCMA\footnote{https://argumentationcompetition.org/2021/},
and random graphs according to \citet{zhou2016spin}.
We establish the optimal solution using
DAGer~\cite{KieselS22,KieselS23}.
We discard any instance not solvable by DAGer.
This results in 18 training instances and 124 test instances.

\paragraph{BDDT}
We consider 69 datasets used in related work on optimal decision trees~\cite{Bessiere09,Olson2017,Narodytska18,Verwer2019,Avellaneda20,SchidlerS24}.
We run the SAT encoding on each instance for ten hours, starting with an initial
bound of 10, as the encoding becomes too large with higher bounds.
We decrease the bound with every solution we find.
The training set consists of 9 instances and and the test set contains 32 instances.

We evaluate a local search function on the training instances
as described in Section~\ref{sec:score}.
We run the local search function with a timeout of 60 seconds but only
count it is a violation of the timeout if no assignment is returned
within 120 seconds.
The one minute timeout is used, as Python is quite slow and shorter timeouts
would make it harder to distinguish the performance of the
local search functions.
The SAT solver is run for at most 120 seconds.
Since
we know that the training instances are solvable
within one minute, 120 seconds allows for enough variance in the runtime.

\subsection{Comprehension}\label{sec:comprehension}
The LLMs do not receive any information on the purpose of the PySAT encoding schemes.
Hence, it is an interesting question if they understand the encodings.
We evaluate this by asking the three LLM models to explain the
PySAT encoding schemes.
All three models perform similarly on this task.
They correctly identify the Graph Coloring and BDDT encoding scheme in detail,
including the concept of the encoding scheme and the semantics of the schemes's variables.
All three models can explain the parts of the DFVS encoding scheme.
However, they fail to identify its overall purpose.

\subsection{Gathering Phase}
The LLM models generate the local search functions
with varying speed during the Gathering Phase.
The difference in performance is also observable in the Refinement Phase, where
it matters less as the evaluation on the training instances takes much longer than
the LLM queries.

The speed of finding the local search functions is mainly determined by two factors:
(i)~how fast the LLM model can come up with new searches despite large contexts,
and,
(ii)~how many repair cycles the LLM model requires to fix code issues, and how long these cycles take (which relates to~(i)).
Generating one local search function can take between a second and several minutes,
mostly depending on how many local search functions we already generated---and are, therefore,
in the context---and how well the LLM model performs regarding~(i).
We observe that GPT o1-mini and Claude Sonnet do not slow down much
with larger contexts, while the speed of GPT 4o significantly decreases
with larger contexts.

The initial version returned by the LLMs is rarely without issues
and repair cycles are required for all LLM models.
However,
Claude Sonnet and GPT 4o often require several iterations and may not be able to fix
their code within ten tries, while GPT o1-mini is able to fix issues in fewer iterations.
Consequently, GPT~o1-mini can generate the 50 searches in the Gathering Phase the fastest,
Claude Sonnet is fast but requires more tokens,
and GPT~4o takes the longest time to complete the Gathering Phase.

\subsection{Refinement Phase}
We pick the top five local search functions from each LLM
model for the Refinement Phase.
This results in 15 \emph{Base} local search versions for each problem.
We refine each of these local searches 19 times, to obtain a total of 20 versions
per local search, which results in 100 local searches per LLM model.
After the first ten refinements (versions 2--11), we re-encourage the LLM to
use the structure of the encoding (versions 12--20).
We refer to versions 2--11 as \emph{Refined} versions and version 12--20 as \emph{Structure}
versions.

\begin{figure}[htb]
    \centering
\cv{
    \begin{subfigure}[b]{\columnwidth}
      \centering
         \includegraphics[scale=0.6, trim=2mm 13.5mm 0 2mm, clip]{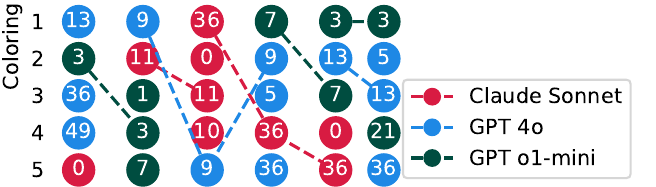}
       \end{subfigure}

       \medskip
    \begin{subfigure}[b]{\columnwidth}
        \centering
         \includegraphics[scale=0.6, trim=2mm 13.5mm 0 2mm, clip]{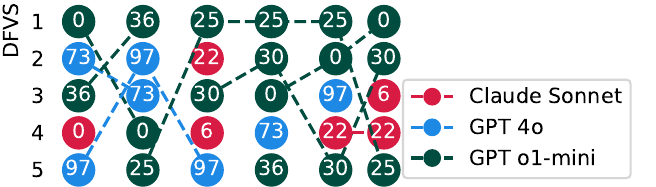}
       \end{subfigure}
       
              \medskip
    \begin{subfigure}[b]{\columnwidth}
        \centering
         \includegraphics[scale=0.6, trim=2mm 3mm 0 2mm, clip]{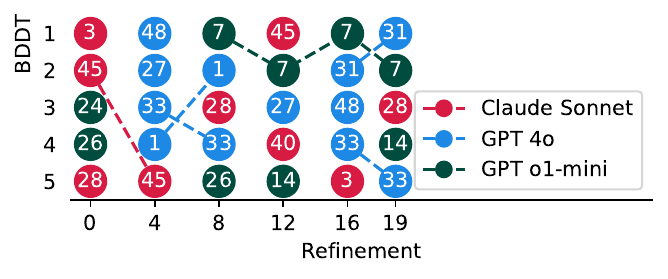}
     \end{subfigure}
}\av{
    \begin{subfigure}[b]{\columnwidth}
      \centering
         \includegraphics[scale=0.7, trim=2mm 13.5mm 0 2mm, clip]{refinement_timeline_coloring}
       \end{subfigure}

       \bigskip
    \begin{subfigure}[b]{\columnwidth}
        \centering
         \includegraphics[scale=0.7, trim=2mm 13.5mm 0 2mm, clip]{refinement_timeline_dfvs}
       \end{subfigure}
       
              \bigskip
    \begin{subfigure}[b]{\columnwidth}
        \centering
         \includegraphics[scale=0.7, trim=2mm 3mm 0 2mm, clip]{refinement_timeline_dt}
     \end{subfigure}
}
     \caption{The top five local search functions for different problems over different
    refinement iterations.}
    \label{fig:refinement-timeline}
\end{figure}

Figure~\ref{fig:refinement-timeline} shows the relative ranking of the
different local search functions at different points in the refinement phase.
This phase indeed achieves significant changes over time, with the ranking constantly changing.
A closer look reveals that the results can indeed change significantly in the course
of a single refinement.
The LLMs also rarely fail to revert a worsening refinement and are,
therefore, able to incorporate our feedback from the training runs.

\subsection{Test Instance Performance}
In this evaluation, the base local search functions and their refinements are run
on the test instances.
Since it is infeasible to evaluate all refinements of all local search functions,
we focus on the Base version, the best Refined version, and the best Structure version.
The best version is determined by the performance on the training instances.
We run the SAT solver without any local search on all instances for one hour as a reference.
We evaluate a local search function on a test instance by running it
with a timeout of 15 minutes and then run the SAT solver using the local
search's assignment for up to one hour.
We use a timeout of 15 minutes and do not decrease the SAT solver timelimit,
as we expect a good native implementation, e.g., in C++, would run orders of magnitudes
faster than the Python prototypes.
The results of the best local search function from each
LLM model are in Table~\ref{tab:eval-summary}.\footnote{The results
for all 15 Base versions are in the supplementary material.}

We are interested in two metrics:
(i)~how many instances are solved by the SAT solver within the timeout when using a local search function,
and
(ii)~how many instances can be solved within the timeout using the local search function
that the SAT solver alone could not solve.
The first metric establishes a general usefulness of the local search function, while
the second metric establishes whether the local search function is useful for hard instances.

\begin{table}[htb]
    \begin{tabular}{@{}l@{} r@{\hspace{5pt}}r@{\hspace{5pt}}r@{\hspace{5pt}}r@{\hspace{5pt}}r@{\hspace{5pt}}r@{}}
        \toprule
        & \multicolumn{2}{c}{Base} & \multicolumn{2}{c}{Refined} & \multicolumn{2}{c}{Structure}\\
        Method & Solved & New & Solved & New & Solved & New\\
        \midrule
\multicolumn{7}{@{}l}{Coloring (38 instances)} \\
        \midrule
SAT&8&-&-&-&-&-\\
Claude Sonnet 36&4&0&5&0&8&3\\
GPT 4o 49&6&1&7&0&5&0\\
GPT o1-mini 7&3&0&12&7&3&0\\
               \midrule
\multicolumn{7}{@{}l}{DFVS (124 instances)} \\
        \midrule
SAT&61&-&-&-&-&-\\
Claude Sonnet 6&38&5&46&6&47&6\\
GPT 4o 73&46&6&44&6&50&8\\
GPT o1-mini 30&60&10&61&12&58&11\\
        \midrule
\multicolumn{7}{@{}l}{BDDT (32 instances)} \\
\midrule
SAT&16&-&-&-&-&-\\
Claude Sonnet 3&14&0&7&0&8&1\\
GPT 4o 31&10&1&7&0&8&0\\
GPT o1-mini 7&10&1&13&1&13&1\\
        \bottomrule
    \end{tabular}
    \centering
    \caption{Results for the best local search functions from each LLM model.
    \emph{Solved} indicates how many instances the method solved
    and \emph{New} how many of those the SAT solver alone could not solve.
    \emph{Base} is the initial version found in the Gathering Phase,
        \emph{Refined} is the best version in the first 10 refinements, and
    \emph{Structure} is the best version in the last 9 refinements.}
    \label{tab:eval-summary}

\end{table}

The results differ for each problem.
It is hard to find hard but solvable instances for graph coloring.
Most instances are either very easy or very hard.
Hence, most of our instances stay unsolved by any method.
Overall, several instances that can be solved without a local search function
are not solved with it.
This indicates that the local search's assignment is not close
enough to a satisfying solution.
This does not necessarily imply that the hybrid approach performs worse,
as the SAT solver alone can also find satisfying assignments by chance.
Hence, local search functions should not be used on instances
that do not require them.
However, the best local search function achieves more solved instances than
the SAT solver alone, showing that the hybrid approach is an improvement.
This local search function would not have been found without the refinement phase,
showing that gathering functions alone is not sufficient for good results.
Overall, already the prototypes created by the LLM look encouraging for hard instances,
we expect an efficient implementation to achieve even better results.

The local searches for DFVS achieve overall relatively better results than
the local searches for graph coloring.
There is also a clear difference in performance between
the different LLM models, with GPT o1-mini
finding the best-performing local search functions, while
Claude Sonnet struggles to find good local search functions.
While the average over all local search functions
looks better for DFVS than for graph coloring, the best function
can only help solve as many instances as the SAT solver alone.
As with graph coloring, several instances go unsolved whenever local search
used, although the SAT solver alone can solve them.
The results show that the best local search function
can aid the SAT solver on several hard instances.
As in the case of graph coloring, refinement is necessary
to find the best performing local search.

BDDT is the problem the LLMs struggled most with.
No matter which local search function is used,
the SAT solver alone can solve more instances.
It is not known how many additional instances can reasonably
be expected to be solvable by a SAT solver.
Hence, the local search functions might perform
much better in case the test instances contained more hard but solvable instances.
One of the best local search functions can aid the SAT solver to
solve instance \emph{objectivity} in 2483 seconds, which was reported by
\citet{SchidlerS24} as not solvable within six hours.
This raises our confidence in the quality of the local search functions.

\begin{figure}
    \centering
\cv{    \includegraphics[scale=0.6]{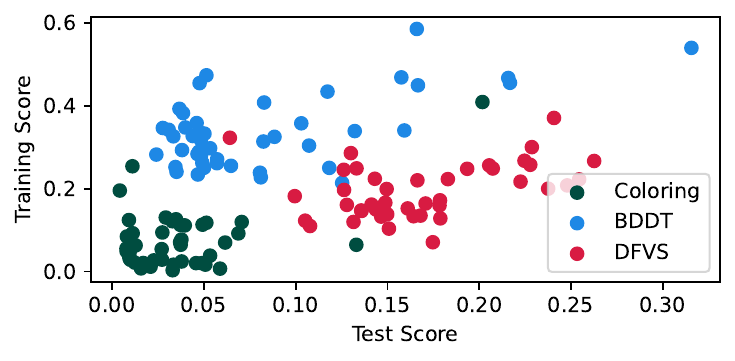}}
\av{    \includegraphics[scale=0.75]{prediction_scatter}}
    \caption{Local search function scores on the training instances
        and test instances.
    Each marker represents one local search function.}
    \label{fig:prediction-scatter}
\end{figure}

\subsection{Discussion}
Our results show that it is indeed possible to automatically generate and
evaluate local search prototypes using LLMs.
In the last part of our evaluation we want to discuss aspects of our
approach that are of interest for adapting our
approach to other problems.

\subsubsection{Test and Training Correlation}
The automatic evaluation relies on some automated way of ranking the different
local search functions.
Since large scale tests for each of the many generated functions are
infeasible, we rely on the training instances as an approximation of performance
on the interesting test instances.
An important question is, how well does the test performance approximate the training
performance?
We explore this using a score that is based on the time it takes
the SAT solver to solve the instance with the assignment provided by the local search.
The score for a local search function on a specific instance is the
fastest time over all local search functions divided by the time for the
scored local search function.
Hence, the faster the result, the closer to one the score, where timeouts are
counted as score 0.
Figure~\ref{fig:prediction-scatter} reports the average score over all
instances.

The correlation is strongest for DFVS, becomes weaker for BDDT,
and is almost non-present for coloring.
DFVS has the most training instances and the training
instances come from several different sources, giving
the training set diversity.
In contrast, coloring instances tend to be either easy or hard
and few instances fit into the 10 to 60  second range.
This leads to many test instance from the same source and
low diversity.
BDDT also has fewer test instance than DFVS, but the
datasets are very different from each other.

The selection of training instances is, therefore, very
important for finding the best performing local
search functions.
Nonetheless, the coloring results show that our approach
can still pick out very good local search functions,
which is also indicated by the very high scoring
marks in Figure~\ref{fig:prediction-scatter}.

\subsubsection{Code Diversity}
We manually reviewed the code generated by the LLMs
in an effort to judge how much the code varies,
as well as the overall quality of the code.
Due to the large number of generated local search functions---over
900---we cannot review all the code in detail.
Hence, we focus on trends within the code.

The LLMs manage to create prototypes and these prototypes are rarely
performance-optimized.
An example is the scoring function present in
all of the local search functions we reviewed.
The scoring function determines how good an assignment is and
is often implemented by iterating over all clauses
and checking if they are satisfied.
The scoring function is then called whenever
the assignment changes, or even for each considered change to the assignment.
In case of methods like WalkSAT that flip one variable
per iteration, this is a major performance bottleneck.
We address this specific issue via prompt and the LLMs can often
fix it, but performance bottlenecks that are specific to single
local search functions cannot be addressed via generic prompt.
Hence, we expect that the local search functions can be improved
even more with a performance-oriented implementation.
Further, potential issues should be assessed before the Gathering Phase
by generating a small set of initial local search functions and manually reviewing them.
These issues can then be addressed directly, as
with the scoring function bottleneck.

The LLMs usually provide their implementation goal in the comments.
According to these comments, the local search functions
implement over 50 different meta-heuristics, like \cv{\enlargethispage{-5mm}}

\begin{quote}
genetic algorithms, tabu search,
simulated annealing, harmony search, ant colony optimization,
agent-based optimization, cellular automaton based optimization, swarm optimization,
great deluge, firefly, bee colony optimization, multi-agent-based optimization,
market-based optimization, quantum-based binary optimization.
\end{quote}
Further, every LLM model also tries WalkSAT in some iteration for every problem.
Interestingly, Claude Sonnet often tries to combine different meta-heuristics, while
the GPT models try to implement a single approach.
This single approach closely resembles pseudocode in local search functions
generated by GPT 4o, while those generated by GPT o1-mini have
clear adaptations due to our prompts.
These adaptations are also observable in the Claude Sonnet generated functions.
This great variety is encouraging for sampling promising prototypes.
Unfortunately, the comments often don't match the semantics of the code,
making a manual review necessary to verify what the code actually does.

Issues are usually filtered away due to bad performance on the training set.
Surprisingly, even wrong implementations can sometimes deliver good results.
Hence, we found several such issues in even the top-performing local search functions.
Consequently, there is room for even better performance when
efficiently implementing the prototypes, but a proper analysis of the
code is necessary beforehand.

\subsubsection{Analysis of Generated Search Strategies}
In this last part, we analyze the code of the best local search functions
and report the most interesting ideas.
For a variable flip, we define its \emph{conflict score} as $S-U$, where $S$ is the number of unsatisfied clauses that become satisfied and $U$ is the number of satisfied clauses that become unsatisfied. Higher scores indicate better flips.

\paragraph{Graph Coloring}
The graph coloring encoding is comparatively simple.
Consequently, the LLMs often generate local search functions for graph coloring
that run on the graph, and only when returning, convert the best coloring
into a variable assignment.
This works well and the best function we found works with this principle.
The best local search function that uses the SAT encoding implements
tabu search and in each iteration performs the best---according to the conflict score---variable
flip among 20 random variables.
Whenever the improvements stagnate, the search uses the 20 random variables whose
flips lead to an improvement most often.
The initial assignment is created by assigning colors to nodes in order if decreasing
node degree, choosing the color that causes the fewest monochromatic edges.

\paragraph{DFVS}
The DFVS PySAT encoding scheme is an interesting case.
As described in Section~\ref{sec:comprehension}, this scheme is the only one
that the LLMs do not fully comprehend.
This leads to strange effects, where several local search functions are able
to extract the upper bound $d$ from the encoding, but misunderstand the polarity
of the variables inside the cardinality constraint.
This leads to the local search function seemingly
excluding d many nodes from the graph, while actually excluding all but d many nodes.

The best local search function is a straightforward WalkSAT implementation.
Other good local search functions use greedy heuristics based on
node degree or node centrality to pick the best variable flips.
While there are many well-performing local search functions,
there is a lack of problem specific functionality, and where
it is present, it is wrong, as with the polarity issue stated above.
Overall, the lesson-learned from this PySAT case, is the
importance of the LLM model's understanding of the encoding scheme.

\paragraph{BDDT}
The best local search functions for BDDT are highly adapted to the PySAT encoding scheme.
The best function for BDDT is generated by Claude Sonnet.
It correctly identifies the variables that must be assigned, and ignores
those variables that are implied by unit propagation.
The remaining variables are separated in levels, corresponding to the
depth of the respective node in the tree.
During initialization, the function assigns each node a random feature and picks
a threshold from the middle of the dataset.
This is reflected in the assignment such that all corresponding constraints are
satisfied.
The search is performed level by level.
Whenever the search on one level stagnates, it moves to the next level,
cycling back to the root level if necessary.
The search itself is performed by first picking a random unsatisfied clause that
contains a variable from the current level and then picks the variable with the
highest conflict score
from the clause .

Another local search method separates the variables in different layers, based on
the features they encode, thresholds, or classes.
The search then
adapts the assignment layer by layer, fixing the
assignment of variables that occur together with the most variables outside
the layer first.
This process is repeated and whenever search stagnates,
the order of layers is changed.

Overall, the quality and specificity of the local search functions for
BDDT shows how well the LLMs can adapt the code to even complicated 
encoding schemes, as long as the LLM model understands them.

\section{Conclusion}
We demonstrated that it is possible to automatically generate
effective local search algorithms by having LLMs analyze SAT encoding
schemes. Our key innovation lies in targeting the encoding methodology
rather than specific problem instances, allowing the generated
strategies to work across all problems sharing the same encoding
pattern. This scheme-centric approach produced diverse search
strategies whose performance correlated with the LLMs' comprehension
of the underlying encoding structures.

While we focused on LLM's,
exploring distilled variants could make the generation process more
computationally feasible. Our diversity measures in the exploration
phase could also be enhanced by leveraging model embeddings to
quantify the distinctness of generated strategies better. Finally,
while computationally more demanding, the next generation of reasoning
models could enable deeper encoding comprehension and more
sophisticated search strategies, further expanding the possibilities
of automated algorithm generation.

\bibliographystyle{named_doi}

\bibliography{main}

\begin{thebibliography}{}

\bibitem[\protect\citeauthoryear{Avellaneda}{2020}]{Avellaneda20}
Florent Avellaneda.
\newblock Efficient inference of optimal decision trees.
\newblock In {\em Proceedings of {AAAI} 2020}. AAAI Press, 2020.

\bibitem[\protect\citeauthoryear{Balint and Manthey}{2013}]{BalintM13}
Adrian Balint and Norbert Manthey.
\newblock Boosting the performance of {SLS} and {CDCL} solvers by preprocessor
  tuning.
\newblock In {\em POS@SAT}, volume~29 of {\em EPiC Series in Computing}, pages
  1--14. EasyChair, 2013.
\newblock DOI: \url{https://doi.org/10.29007/28WW}

\bibitem[\protect\citeauthoryear{Balint and Sch{\"{o}}ning}{2012}]{BalintS12}
Adrian Balint and Uwe Sch{\"{o}}ning.
\newblock Choosing probability distributions for stochastic local search and
  the role of make versus break.
\newblock In {\em Theory and Applications of Satisfiability Testing - {SAT}
  2012 - 15th International Conference, Trento, Italy, June 17-20, 2012.
  Proceedings}, volume 7317 of {\em Lecture Notes in Computer Science}, pages
  16--29. Springer, 2012.
\newblock DOI: \url{https://doi.org/10.1007/978-3-642-31612-8\_3}

\bibitem[\protect\citeauthoryear{Bessiere \bgroup \em et al.\egroup
  }{2009}]{Bessiere09}
Christian Bessiere, Emmanuel Hebrard, and Barry O'Sullivan.
\newblock Minimising decision tree size as combinatorial optimisation.
\newblock In {\em Proceedings of {CP} 2009}, pages 173--187, Berlin,
  Heidelberg, 2009. Springer Berlin Heidelberg.

\bibitem[\protect\citeauthoryear{Biere \bgroup \em et al.\egroup
  }{2020}]{BiereFHH20}
Armin Biere, Katalin Fazekas, Mathias Fleury, and Maximilian Heisinger.
\newblock {CaDiCaL}, {Kissat}, {Paracooba}, {Plingeling}, and {Treengeling}
  entering the {SAT} competition 2020.
\newblock In {\em Proceedings of {SAT} Competition 2020: Solver and Benchmark
  Descriptions}, pages 51--53, 2020.

\bibitem[\protect\citeauthoryear{Biere \bgroup \em et al.\egroup
  }{2024}]{Biere24}
Armin Biere, Tobias Faller, Katalin Fazekas, Mathias Fleury, Nils Froleyks, and
  Florian Pollitt.
\newblock Cadical 2.0.
\newblock In {\em Computer Aided Verification - 36th International Conference,
  {CAV} 2024, Montreal, QC, Canada, July 24-27, 2024, Proceedings, Part {I}},
  volume 14681 of {\em Lecture Notes in Computer Science}, pages 133--152.
  Springer, 2024.
\newblock DOI: \url{https://doi.org/10.1007/978-3-031-65627-9\_7}

\bibitem[\protect\citeauthoryear{Biere}{2014}]{Biere14}
Armin Biere.
\newblock Yet another local search solver and {Lingeling} and friends entering
  the {SAT Competition 2014}.
\newblock In {\em Proc.~of {SAT Competition} 2014 -- Solver and Benchmark
  Descriptions}, volume B-2014-2 of {\em Department of Computer Science Series
  of Publications B}, pages 39--40. University of Helsinki, 2014.

\bibitem[\protect\citeauthoryear{Biere}{2019}]{Biere19}
Armin Biere.
\newblock {CaDiCaL} at the {SAT} race 2019.
\newblock In {\em Proceedings of {SAT} Race 2019: Solver and Benchmark
  Descriptions}, pages 8--9, 2019.

\bibitem[\protect\citeauthoryear{Br\'{e}laz}{1979}]{Brelaz79}
Daniel Br\'{e}laz.
\newblock New methods to color the vertices of a graph.
\newblock {\em Commun. ACM}, 22(4):251–256, apr 1979.
\newblock URL: \url{https://dl.acm.org/doi/10.1145/359094.359101}

\bibitem[\protect\citeauthoryear{Cai and Zhang}{2021}]{CaiZ21}
Shaowei Cai and Xindi Zhang.
\newblock Deep cooperation of {CDCL} and local search for {SAT}.
\newblock In {\em Theory and Applications of Satisfiability Testing - {SAT}
  2021 - 24th International Conference, Barcelona, Spain, July 5-9, 2021,
  Proceedings}, volume 12831 of {\em Lecture Notes in Computer Science}, pages
  64--81. Springer, 2021.
\newblock DOI: \url{https://doi.org/10.1007/978-3-030-80223-3\_6}

\bibitem[\protect\citeauthoryear{Cai \bgroup \em et al.\egroup
  }{2022}]{CaiZFB22}
Shaowei Cai, Xindi Zhang, Mathias Fleury, and Armin Biere.
\newblock Better decision heuristics in {CDCL} through local search and target
  phases.
\newblock {\em J. Artif. Intell. Res.}, 74:1515--1563, 2022.
\newblock DOI: \url{https://doi.org/10.1613/JAIR.1.13666}

\bibitem[\protect\citeauthoryear{Cummins \bgroup \em et al.\egroup
  }{2023}]{CumminsSGELRGGHSL23}
Chris Cummins, Volker Seeker, Dejan Grubisic, Mostafa Elhoushi, Youwei Liang,
  Baptiste Rozi{\`{e}}re, Jonas Gehring, Fabian Gloeckle, Kim~M. Hazelwood,
  Gabriel Synnaeve, and Hugh Leather.
\newblock Large language models for compiler optimization.
\newblock {\em CoRR}, abs/2309.07062, 2023.
\newblock DOI: \url{https://doi.org/10.48550/ARXIV.2309.07062}

\bibitem[\protect\citeauthoryear{Fichte \bgroup \em et al.\egroup
  }{2017}]{Fichte17}
Johannes~Klaus Fichte, Neha Lodha, and Stefan Szeider.
\newblock {SAT}-based local improvement for finding tree decompositions of
  small width.
\newblock In {\em Theory and Applications of Satisfiability Testing - {SAT}
  2017 - 20th International Conference, Melbourne, VIC, Australia, August 28 -
  September 1, 2017, Proceedings}, volume 10491 of {\em Lecture Notes in
  Computer Science}, pages 401--411. Springer, 2017.
\newblock DOI: \url{https://doi.org/10.1007/978-3-319-66263-3\_25}

\bibitem[\protect\citeauthoryear{Fichte \bgroup \em et al.\egroup
  }{2023}]{FichteBHS23}
Johannes~Klaus Fichte, Daniel~Le Berre, Markus Hecher, and Stefan Szeider.
\newblock The silent (r)evolution of {SAT}.
\newblock {\em Commun. {ACM}}, 66(6):64--72, 2023.
\newblock DOI: \url{https://doi.org/10.1145/3560469}

\bibitem[\protect\citeauthoryear{Gelder}{2008}]{Gelder08}
Allen~Van Gelder.
\newblock Another look at graph coloring via propositional satisfiability.
\newblock {\em Discret. Appl. Math.}, 156(2):230--243, 2008.
\newblock DOI: \url{https://doi.org/10.1016/J.DAM.2006.07.016}

\bibitem[\protect\citeauthoryear{Gro{\ss}mann \bgroup \em et al.\egroup
  }{2022}]{PACE22}
Ernestine Gro{\ss}mann, Tobias Heuer, Christian Schulz, and Darren Strash.
\newblock The {PACE} 2022 parameterized algorithms and computational
  experiments challenge: Directed feedback vertex set.
\newblock In {\em Proceedings of {IPEC} 2022}, volume 249 of {\em LIPIcs},
  pages 26:1--26:18. Schloss Dagstuhl - Leibniz-Zentrum f{\"{u}}r Informatik,
  2022.
\newblock DOI: \url{https://doi.org/10.4230/LIPICS.IPEC.2022.26}

\bibitem[\protect\citeauthoryear{Hoos}{2002}]{Hoos02}
Holger~H. Hoos.
\newblock An adaptive noise mechanism for walksat.
\newblock In {\em Proceedings of the Eighteenth National Conference on
  Artificial Intelligence and Fourteenth Conference on Innovative Applications
  of Artificial Intelligence, July 28 - August 1, 2002, Edmonton, Alberta,
  Canada}, pages 655--660. {AAAI} Press / The {MIT} Press, 2002.
\newblock URL: \url{http://www.aaai.org/Library/AAAI/2002/aaai02-098.php}

\bibitem[\protect\citeauthoryear{Ignatiev \bgroup \em et al.\egroup
  }{2018}]{Ignatiev18}
Alexey Ignatiev, Antonio Morgado, and Joao Marques{-}Silva.
\newblock {PySAT:} {A} {Python} toolkit for prototyping with {SAT} oracles.
\newblock In {\em SAT}, pages 428--437, 2018.
\newblock DOI: \url{https://doi.org/10.1007/978-3-319-94144-8_26}

\bibitem[\protect\citeauthoryear{Ignatiev \bgroup \em et al.\egroup
  }{2024}]{IgnatievTK24}
Alexey Ignatiev, Zi~Li Tan, and Christos Karamanos.
\newblock Towards universally accessible {SAT} technology.
\newblock In {\em 27th International Conference on Theory and Applications of
  Satisfiability Testing, {SAT} 2024, August 21-24, 2024, Pune, India}, volume
  305 of {\em LIPIcs}, pages 16:1--16:11. Schloss Dagstuhl - Leibniz-Zentrum
  f{\"{u}}r Informatik, 2024.
\newblock DOI: \url{https://doi.org/10.4230/LIPICS.SAT.2024.16}

\bibitem[\protect\citeauthoryear{Janota \bgroup \em et al.\egroup
  }{2017}]{JanotaGM17}
Mikol{\'{a}}s Janota, Radu Grigore, and Vasco Manquinho.
\newblock On the quest for an acyclic graph.
\newblock In {\em Proceedings of the 24th {RCRA} International Workshop on
  Experimental Evaluation of Algorithms for Solving Problems with Combinatorial
  Explosion 2017, Bari, Italy, November 14-15, 2017}, volume 2011 of {\em
  {CEUR} Workshop Proceedings}, pages 33--44. CEUR-WS.org, 2017.
\newblock URL: \url{https://ceur-ws.org/Vol-2011/paper4.pdf}

\bibitem[\protect\citeauthoryear{Jiang \bgroup \em et al.\egroup
  }{2024}]{JiangWSKK2024}
Juyong Jiang, Fan Wang, Jiasi Shen, Sungju Kim, and Sunghun Kim.
\newblock A survey on large language models for code generation.
\newblock {\em CoRR}, abs/2406.00515, 2024.
\newblock DOI: \url{https://doi.org/10.48550/ARXIV.2406.00515}

\bibitem[\protect\citeauthoryear{Karp}{1972}]{karp1972reducibility}
Richard~M. Karp.
\newblock Reducibility among combinatorial problems.
\newblock In {\em Complexity of Computer Computations 1972}, The {IBM} Research
  Symposia Series, pages 85--103. Plenum Press, New York, 1972.
\newblock DOI: \url{https://doi.org/10.1007/978-1-4684-2001-2\_9}

\bibitem[\protect\citeauthoryear{Kiesel and Schidler}{2022}]{KieselS22}
Rafael Kiesel and Andr{\'{e}} Schidler.
\newblock {PACE} solver description: Dager - cutting out cycles with maxsat.
\newblock In {\em Proceedings of {IPEC} 2022}, volume 249 of {\em LIPIcs},
  pages 32:1--32:4. Schloss Dagstuhl - Leibniz-Zentrum f{\"{u}}r Informatik,
  2022.
\newblock DOI: \url{https://doi.org/10.4230/LIPICS.IPEC.2022.32}

\bibitem[\protect\citeauthoryear{Kiesel and Schidler}{2023}]{KieselS23}
Rafael Kiesel and Andr{\'{e}} Schidler.
\newblock A dynamic {MaxSAT}-based approach to directed feedback vertex sets.
\newblock In {\em Proceedings of the Symposium on Algorithm Engineering and
  Experiments, {ALENEX} 2023, Florence, Italy, January 22-23, 2023}, pages
  39--52. {SIAM}, 2023.
\newblock DOI: \url{https://doi.org/10.1137/1.9781611977561.CH4}

\bibitem[\protect\citeauthoryear{Li and Li}{2012}]{LiL12}
Chu~Min Li and Yu~Li.
\newblock Satisfying versus falsifying in local search for satisfiability -
  (poster presentation).
\newblock In {\em Theory and Applications of Satisfiability Testing - {SAT}
  2012 - 15th International Conference, Trento, Italy, June 17-20, 2012.
  Proceedings}, volume 7317 of {\em Lecture Notes in Computer Science}, pages
  477--478. Springer, 2012.
\newblock DOI: \url{https://doi.org/10.1007/978-3-642-31612-8\_43}

\bibitem[\protect\citeauthoryear{Li \bgroup \em et al.\egroup
  }{2022}]{LiCCKSLEKGLHCMBCHWGCMCMRKFKV22}
Yujia Li, David~H. Choi, Junyoung Chung, Nate Kushman, Julian Schrittwieser,
  R{\'{e}}mi Leblond, Tom Eccles, James Keeling, Felix Gimeno, Agustin~Dal
  Lago, Thomas Hubert, Peter Choy, Cyprien de~Masson~d'Autume, Igor Babuschkin,
  Xinyun Chen, Po{-}Sen Huang, Johannes Welbl, Sven Gowal, Alexey Cherepanov,
  James Molloy, Daniel~J. Mankowitz, Esme~Sutherland Robson, Pushmeet Kohli,
  Nando de~Freitas, Koray Kavukcuoglu, and Oriol Vinyals.
\newblock Competition-level code generation with alphacode.
\newblock {\em CoRR}, abs/2203.07814, 2022.
\newblock DOI: \url{https://doi.org/10.48550/ARXIV.2203.07814}

\bibitem[\protect\citeauthoryear{Marques{-}Silva \bgroup \em et al.\egroup
  }{2021}]{DBLP:series/faia/0001LM21}
Jo{\~{a}}o Marques{-}Silva, In{\^{e}}s Lynce, and Sharad Malik.
\newblock Conflict-driven clause learning {SAT} solvers.
\newblock In {\em Handbook of Satisfiability - Second Edition}, volume 336 of
  {\em Frontiers in Artificial Intelligence and Applications}, pages 133--182.
  {IOS} Press, 2021.
\newblock DOI: \url{https://doi.org/10.3233/FAIA200987}

\bibitem[\protect\citeauthoryear{McAllester \bgroup \em et al.\egroup
  }{1997}]{McAllesterSK97}
David~A. McAllester, Bart Selman, and Henry~A. Kautz.
\newblock Evidence for invariants in local search.
\newblock In {\em Proceedings of the Fourteenth National Conference on
  Artificial Intelligence and Ninth Innovative Applications of Artificial
  Intelligence Conference, {AAAI} 97, {IAAI} 97, July 27-31, 1997, Providence,
  Rhode Island, {USA}}, pages 321--326. {AAAI} Press / The {MIT} Press, 1997.
\newblock URL: \url{http://www.aaai.org/Library/AAAI/1997/aaai97-050.php}

\bibitem[\protect\citeauthoryear{Narodytska \bgroup \em et al.\egroup
  }{2018}]{Narodytska18}
Nina Narodytska, Alexey Ignatiev, Filipe Pereira, and Joao Marques-Silva.
\newblock Learning optimal decision trees with {SAT}.
\newblock In {\em Proceedings of {IJCAI} 2018}, pages 1362--1368. ijcai.org, 7
  2018.
\newblock DOI: \url{https://doi.org/10.24963/ijcai.2018/189}

\bibitem[\protect\citeauthoryear{Nijkamp \bgroup \em et al.\egroup
  }{2023}]{NijkampPHTWZSX23}
Erik Nijkamp, Bo~Pang, Hiroaki Hayashi, Lifu Tu, Huan Wang, Yingbo Zhou, Silvio
  Savarese, and Caiming Xiong.
\newblock Codegen: An open large language model for code with multi-turn
  program synthesis.
\newblock In {\em {ICLR}}. OpenReview.net, 2023.

\bibitem[\protect\citeauthoryear{Olson \bgroup \em et al.\egroup
  }{2017}]{Olson2017}
Randal~S. Olson, William La~Cava, Patryk Orzechowski, Ryan~J. Urbanowicz, and
  Jason~H. Moore.
\newblock {PMLB}: a large benchmark suite for machine learning evaluation and
  comparison.
\newblock {\em BioData Mining}, 10(1):36, Dec 2017.
\newblock DOI: \url{https://doi.org/10.1186/s13040-017-0154-4}

\bibitem[\protect\citeauthoryear{Schidler and Szeider}{2024}]{SchidlerS24}
Andr{\'{e}} Schidler and Stefan Szeider.
\newblock {SAT}-based decision tree learning for large data sets.
\newblock {\em J. Artif. Intell. Res.}, 80:875--918, 2024.
\newblock DOI: \url{https://doi.org/10.1613/JAIR.1.15956}

\bibitem[\protect\citeauthoryear{Schidler and
  Szeider}{2025}]{SchidlerSzeider25}
André Schidler and Stefan Szeider.
\newblock Extracting problem structure with {LLMs} for optimized {SAT} local
  search, January 2025.
\newblock DOI: \url{https://doi.org/10.5281/zenodo.14732109}

\bibitem[\protect\citeauthoryear{Selman \bgroup \em et al.\egroup
  }{1992}]{SelmanLM92}
Bart Selman, Hector~J. Levesque, and David~G. Mitchell.
\newblock A new method for solving hard satisfiability problems.
\newblock In {\em Proceedings of the 10th National Conference on Artificial
  Intelligence, San Jose, CA, USA, July 12-16, 1992}, pages 440--446. {AAAI}
  Press / The {MIT} Press, 1992.
\newblock URL: \url{http://www.aaai.org/Library/AAAI/1992/aaai92-068.php}

\bibitem[\protect\citeauthoryear{Selman \bgroup \em et al.\egroup
  }{1994}]{SelmanKC94}
Bart Selman, Henry~A. Kautz, and Bram Cohen.
\newblock Noise strategies for improving local search.
\newblock In {\em Proceedings of the 12th National Conference on Artificial
  Intelligence, Seattle, WA, USA, July 31 - August 4, 1994, Volume 1}, pages
  337--343. {AAAI} Press / The {MIT} Press, 1994.
\newblock URL: \url{http://www.aaai.org/Library/AAAI/1994/aaai94-051.php}

\bibitem[\protect\citeauthoryear{Selsam and Bj{\o}rner}{2019}]{SelsamB19}
Daniel Selsam and Nikolaj~S. Bj{\o}rner.
\newblock Guiding high-performance {SAT} solvers with unsat-core predictions.
\newblock In Mikol{\'{a}}s Janota and In{\^{e}}s Lynce, editors, {\em Theory
  and Applications of Satisfiability Testing - {SAT} 2019 - 22nd International
  Conference, {SAT} 2019, Lisbon, Portugal, July 9-12, 2019, Proceedings},
  volume 11628 of {\em Lecture Notes in Computer Science}, pages 336--353.
  Springer, 2019.
\newblock DOI: \url{https://doi.org/10.1007/978-3-030-24258-9_24}

\bibitem[\protect\citeauthoryear{Shati \bgroup \em et al.\egroup
  }{2021}]{ShatiCM21}
Pouya Shati, Eldan Cohen, and Sheila~A. McIlraith.
\newblock {SAT}-based approach for learning optimal decision trees with
  non-binary features.
\newblock In {\em {CP}}, volume 210 of {\em LIPIcs}, pages 50:1--50:16. Schloss
  Dagstuhl - Leibniz-Zentrum f{\"{u}}r Informatik, 2021.
\newblock DOI: \url{https://doi.org/10.4230/LIPICS.CP.2021.50}

\bibitem[\protect\citeauthoryear{Silva and Sakallah}{1996}]{SilvaS96}
Jo{\~{a}}o P.~Marques Silva and Karem~A. Sakallah.
\newblock {GRASP} - a new search algorithm for satisfiability.
\newblock In {\em Proceedings of the 1996 {IEEE/ACM} International Conference
  on Computer-Aided Design, {ICCAD} 1996, San Jose, CA, USA, November 10-14,
  1996}, pages 220--227. {IEEE} Computer Society / {ACM}, 1996.
\newblock DOI: \url{https://doi.org/10.1109/ICCAD.1996.569607}

\bibitem[\protect\citeauthoryear{Soos}{2020}]{Soos20}
Mate Soos.
\newblock Cryptominisat 5.6.8.
\newblock In {\em Proceedings of {SAT} Competition 2020: Solver and Benchmark
  Descriptions}, pages 37--38, 2020.

\bibitem[\protect\citeauthoryear{Sun \bgroup \em et al.\egroup
  }{2021}]{SunHZL21}
Wen Sun, Jin{-}Kao Hao, Yuhao Zang, and Xiangjing Lai.
\newblock A solution-driven multilevel approach for graph coloring.
\newblock {\em Appl. Soft Comput.}, 104:107174, 2021.
\newblock DOI: \url{https://doi.org/10.1016/J.ASOC.2021.107174}

\bibitem[\protect\citeauthoryear{Szeider}{2024}]{Szeider24}
Stefan Szeider.
\newblock {MCP}-solver: Integrating language models with constraint programming
  systems.
\newblock {\em CoRR}, abs/2501.00539, 2024.
\newblock DOI: \url{https://doi.org/10.48550/ARXIV.2501.00539}

\bibitem[\protect\citeauthoryear{Verwer and Zhang}{2019}]{Verwer2019}
Sicco Verwer and Yingqian Zhang.
\newblock Learning optimal classification trees using a binary linear program
  formulation.
\newblock In {\em Proceedings of {AAAI} 2019}, pages 1625--1632. {AAAI} Press,
  2019.
\newblock DOI: \url{https://doi.org/10.1609/aaai.v33i01.33011624}

\bibitem[\protect\citeauthoryear{Voboril \bgroup \em et al.\egroup
  }{2024}]{VoborilRS24}
Florentina Voboril, Vaidyanathan~Peruvemba Ramaswamy, and Stefan Szeider.
\newblock Generating streamlining constraints with large language models.
\newblock {\em CoRR}, abs/2408.10268, 2024.
\newblock DOI: \url{https://doi.org/10.48550/ARXIV.2408.10268}

\bibitem[\protect\citeauthoryear{Voboril \bgroup \em et al.\egroup
  }{2025}]{StreamLLM}
Florentina Voboril, Vaidyanathan~Peruvemba Ramaswamy, and Stefan Szeider.
\newblock Realtime generation of streamliners with large language models.
\newblock {\em NSE 2025, the First International Workshop on Neuro-Symbolic
  Software Engineering (May 3, 2025), affiliated with ICSE 2025, the IEEE/ACM
  International Conference on Software Engineering}, 2025.

\bibitem[\protect\citeauthoryear{Yolcu and P{\'{o}}czos}{2019}]{YolcuP19}
Emre Yolcu and Barnab{\'{a}}s P{\'{o}}czos.
\newblock Learning local search heuristics for boolean satisfiability.
\newblock In Hanna~M. Wallach, Hugo Larochelle, Alina Beygelzimer, Florence
  d'Alch{\'{e}}{-}Buc, Emily~B. Fox, and Roman Garnett, editors, {\em Advances
  in Neural Information Processing Systems 32: Annual Conference on Neural
  Information Processing Systems 2019, NeurIPS 2019, December 8-14, 2019,
  Vancouver, BC, Canada}, pages 7990--8001, 2019.
\newblock URL:
  \url{https://proceedings.neurips.cc/paper/2019/hash/12e59a33dea1bf0630f46edfe13d6ea2-Abstract.html}

\bibitem[\protect\citeauthoryear{Zhou}{2016}]{zhou2016spin}
Hai-Jun Zhou.
\newblock A spin glass approach to the directed feedback vertex set problem.
\newblock {\em Journal of Statistical Mechanics: Theory and Experiment},
  2016(7):073303, 2016.
\newblock URL: \url{https://doi.org/10.1088/1742-5468/2016/07/073303}

\end{thebibliography}
 \end{document}